\documentclass[journal]{IEEEtran}
\usepackage{graphicx}
\usepackage{makecell}
\usepackage{epstopdf}
\usepackage[colorlinks,urlcolor=red]{hyperref}
\usepackage{amsfonts}
\usepackage{color}
\usepackage{multirow}
\usepackage{setspace}
\usepackage{booktabs}
\usepackage{enumerate}
\usepackage{algorithmic}
\usepackage{amsmath}

\usepackage{subfigure}
\usepackage{bbm}

\usepackage[linesnumbered,ruled,vlined]{algorithm2e}

\begin{document}
%
\title{View-Invariant Skeleton-based Action Recognition via Global-Local Contrastive Learning}
%
%
%
\author{ Cunling~Bian,~\thanks{C.~Bian, W.~Feng$^\dagger$, and S.~Wang$^\dagger$ are with the School of Computer Science and Technology, College of Intelligence and Computing, Tianjin University, Tianjin 300350, China, and also with the Key Research Center for Surface Monitoring and Analysis of Cultural Relics (SMARC), State Administration of Cultural Heritage, China,  and also with the Tianjin Key
Laboratory of Cognitive Computing and Application, Tianjin University,
		Tianjin 300350, China. Email:~clbian@tju.edu.cn, wfeng@ieee.org, songwang@cec.sc.edu.}Wei~Feng$^\dagger$~\IEEEmembership{Member,~IEEE,}~Fanbo~Meng~and~Song~Wang$^\dagger$~\IEEEmembership{Senior Member,~IEEE}
	\thanks{Fanbo~Meng is with the Institute of International Engineering, Tianjin University, Tianjin 300350, China. Email: FbM1999@tju.edu.cn.}	\thanks{S. Wang$^\dagger$ is also with Department of Computer Science and Engineering, University of South Carolina, Columbia, SC 29208, USA.}\thanks{(Co-corresponding authors:~W.~Feng and S.~Wang)}\thanks{This work was  supported,  in  part,  by  the  NSFC  under  Grants  U1803264, 61672376, 61671325.}}


%
%

\markboth{IEEE Transactions on Circuits and Systems for Video Technology, In Submission}%
{Shell \MakeLowercase{\textit{et al.}}: Bare Demo of IEEEtran.cls for IEEE Journals}
%



\maketitle

\begin{abstract}
Skeleton-based human action recognition has been drawing more interest recently due to its low sensitivity to appearance changes and the accessibility of more skeleton data.  However, even the 3D skeletons captured in practice are still sensitive to the viewpoint and direction gave the occlusion of different human-body joints and the errors in human joint localization. Such view variance of skeleton data may significantly affect the performance of action recognition. To address this issue, we propose in this paper a new view-invariant representation learning approach, without any manual action labeling, for skeleton-based human action recognition. Specifically, we leverage the multi-view skeleton data simultaneously taken for the same person in the network training, by maximizing the mutual information between the representations extracted from different views, and then propose a global-local contrastive loss to model the multi-scale co-occurrence relationships in both spatial and temporal domains. Extensive experimental results show that the proposed method is robust to the view difference of the input skeleton data and significantly boosts the performance of unsupervised skeleton-based human action methods, resulting in new state-of-the-art accuracies on two challenging multi-view benchmarks of PKUMMD and NTU RGB+D.
\end{abstract}


%
\IEEEpeerreviewmaketitle

\section{Introduction}
\label{sec:intro}
%
%
%
%
%
%

\IEEEPARstart{H}{uman} action recognition plays an important role in video surveillance, human-machine interaction, and sport video analysis~\cite{herath2017going}.  
Different modality information, such as appearance, depth, optical flows, and body skeletons~\cite{bhardwaj2019efficient} has been used for human action recognition.
Among them, the skeleton consists of compact positions of major body joints~\cite{zhang2020semantics} and can provide highly effective information on human motion underlying different actions~\cite{johansson1973visual,bian2021structural}. 
Skeleton-based action recognition is robust to appearance inconsistencies, different environments, and varying illuminations and is getting more accessible with the rapid development of sensor technology for capturing the skeleton.

\begin{figure}
	\centering
	\includegraphics[trim={0 0 0 0},clip,scale=0.85]{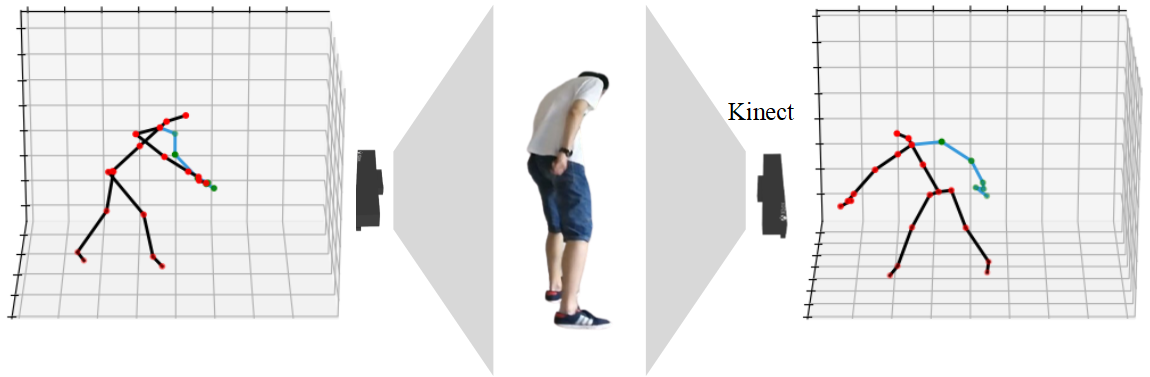}
	\caption{An illustration of view variance of skeleton data: 3d skeleton data simultaneously captured for a same person, but from different views, are usually different due to altered reference coordinates, different occluded joints, or inaccurate human pose estimation.}
	\label{skeleton}
	\vspace{0pt}
\end{figure}

3D skeletons simultaneously captured for the same person from different views are usually different~\cite{zhang2019view}, as shown in Figure~\ref{skeleton}, even if we try to transform them to the same coordinates. There are many reasons accounting for this phenomenon, such as altered reference coordinates, different occluded joints in different views, and inaccurate human pose estimation.
In practice, the skeleton data used for action recognition may be captured from different and even time-varying views, and such view variance can easily lead to incorrect feature representation and
recognition~\cite{zhang2019view,nie2019view}. Nowadays, view-invariant action recognition is still a challenging problem.

A commonly used strategy to improve view invariance in skeleton-based action recognition is to perform frame-level or sequence-level pre-processing of skeleton transformation~\cite{liu2016spatio,song2017end,liu2017global,zhang2019view}. Nevertheless, the frame-level pre-processing transforms the skeleton to the body center with the upper-body orientation aligned and results in partial loss
of relative motion information. To further preserve the motion information, sequence-level pre-processing performs the same transformation on all frames with the same parameters derived in the first frame. However, since the human body is non-rigid, the definition of body plane by the specific joints is not always suitable for the purpose of orientation alignment. Furthermore, it is almost impossible to eliminate the structural differences in different viewpoints through these transformations. Manually designing a view-invariant representation of the action is another approach, such as the displacement of joints within one frame or between frames~\cite{yang2012eigenjoints}, the histogram of joint orientations~\cite{xia2012view}, and some higher-level features like Lie group~\cite{vemulapalli2014human} and the covariant matrix of joints~\cite{hussein2013human}, but they could only deal with small
view changes. While the deep learning-based methods vastly outperform the traditional hand-crafted feature-based methods, most of them just rely on training on a large number of labeled samples taken from various views and cannot achieve actual view invariance in the underlying representation learning.

Most of the state-of-the-art methods for skeleton-based action recognition use supervised deep learning, which requires large-scale annotated data samples for training~\cite{liu2020disentangling,Cheng2020Skeleton}.
To address this problem, several recent studies attempt to leverage unsupervised learning for skeleton-based action recognition~\cite{zheng2018unsupervised,lin2020ms2l,su2020predict}.
In these studies, deep representations are learned for skeleton data sequences in terms of tasks like human motion prediction or regeneration, without using any action labels for supervision. 
For algorithm evaluation, a simple linear classifier is finally trained for action recognition based on both the learned representations and action labels of the training data.
At present, there is still a relatively obvious performance gap between the supervised and unsupervised methods for skeleton-based action recognition. One primary cause lies in that the existing unsupervised skeleton representation learning cannot effectively exclude all the irrelevant factors for action recognition, such as view variation, skeleton deformation, and noise. 
For this reason, we propose a new approach to enhancing representation learning by tackling view variation in skeleton-based action recognition
without using any manual action labels. Since the training data are unlabeled and have been already formated and related in data collection and prepossessing,
we follow previous studies~\cite{li2018unsupervised,ji2021view}, where a surrogate task is designed to exploit the inherent structure of unlabeled multi-view data for representation learning, by calling our proposed method unsupervised in this paper. 

In this paper, we adopt contrastive representation learning to enhance the view-invariant unsupervised skeleton-based action recognition. More specifically, we develop a multi-view spatial-temporal graph (ST-Graph) contrastive representation learning  (CRL) approach, in which the training loss function is defined to maximize the mutual information between features learned from skeletons that are simultaneously captured for the same person from different views. Such multi-view representations of the same person are pulled closer to each other in the embedding space through network training.
Furthermore, the proposed loss takes the form of a global-local contrastive one, which can also model the multi-scale co-occurrence relationships between the spatial and temporal domains. 
In the testing stage, just like in previous works we only take one skeleton data sequence captured from an unknown view as the input of the network for skeleton-based action recognition. 
We conduct comprehensive evaluation and analysis in our experiments to demonstrate that the proposed method can better learn view-invariant representations for improving the performance of skeleton-based human action recognition. The proposed method achieves a new state-of-the-art performance of unsupervised skeleton-based action recognition on two widely used multi-view benchmarks under the linear evaluation protocol.

The main contributions of this paper are as follows:
\begin{itemize}
	\item[-] A contrastive learning framework for explicit learning of view-invariant representations for skeleton-based action recognition is proposed. 
	
	\item[-] We introduce a local-global spatial-temporal graph contrastive loss, combined with task uncertainty, to model the multi-scale co-occurrence relationship between spatial and temporal domains.
	
	\item[-] Compared with existing methods that do not use ground-truth action labels in training, the proposed algorithm significantly boosts the performance on two widely used benchmarks of PKUMMD and NTU RGB+D,  
\end{itemize}

The remainder of the paper is organized as follows. Section 2 gives a brief review of the related work on skeleton-based action recognition and contrastive learning. In Section 3, we describe our proposed multi-view contrastive representation learning approach. Section 4 describes the benchmark datasets and experimental setting and reports the experiment results, followed by a brief conclusion in Section 5.

\section{Related Work}

\subsection{Skeleton-based Action Recognition}

Skeleton-based action recognition is a very active and burgeoning area of research, due to its effective representation of motion dynamics. Much of the traditional skeleton-based action recognition work focuses on designing effective handcrafted features, especially the joint or body part-based features~\cite{vemulapalli2014human,yang2012eigenjoints,xia2012view,hussein2013human}. New methods have recently emerged in the literature to address the skeleton-based action representation with deep learning, including Recurrent Neural Network (RNN), Convolutional Neural Networks (CNN), and Graph Convolutional Network (GCN). Most of them aim to find more effective ways to model temporal and spatial information of skeleton sequences. The structure of RNN is suitable for processing sequential data and prior works have shown that RNN is especially good for handling varying-length skeleton sequences~\cite{wang2017modeling}. To extract discriminative spatial and temporal features of different actions, Song et al.~\cite{song2017end} propose a spatial and temporal attention module to assign different importance to each joint and frame within a sequence on top of RNN. CNN has the intrinsic ability to learn structural information from 2D or 3D grids, and it has also been used to encode skeleton sequences as pseudo-images for spatial-temporal representation learning~\cite{ke2017new}. Liu et al.~\cite{liu2017enhanced} firstly transform skeleton sequence into a series of color images, and then enhance visual and motion local patterns through mathematical morphology, finally propose a multi-stream CNN-based model to extract and fuse deep features from the enhanced-color images. GCN is the generalization of CNN to graphs and it can well represent the joint-based skeleton data. Therefore the use of GCN can automatically capture the patterns embedded in the spatial configuration of the joints as well as their temporal dynamics~\cite{yan2018spatial,liu2020disentangling,Cheng2020Skeleton}.
Cheng et al.~\cite{Cheng2020Skeleton} take novel shift graph operations and lightweight point-wise convolutions to replace regular graph convolutions. This way it reduces computation cost and provides flexible receptive fields for both the spatial graph and the temporal graph.

To avoid the laborious labeling of large-scale skeleton data, unsupervised skeleton-based action recognition has been studied by many researchers~\cite{lin2020ms2l,su2020predict}. 
It performs the feature learning by an encoder-decoder structure, the input of which is a masked or original skeleton sequence, and the goal of training is to reconstruct the skeleton sequences from the encoded features. For the same reason, we focus on enhancing the view-invariant representative learning for skeleton-based action recognition without any manual action labeling using the GCN-based network in this paper.

\subsection{Contrastive Learning}
Contrastive learning aims to pull together an anchor and a “positive” sample in embedding space while pushing apart the anchor from many “negative” samples~\cite{khosla2020supervised}. 
Therefore, contrastive losses are adopted to learn effective representations for pretext tasks in an unsupervised fashion. 
Closely related to contrastive learning is the family of losses based on metric distance learning or triplets that depend on class labels to supervise the choice of positive and negative pairs~\cite{schroff2015facenet}. The key distinction between triplet losses and contrastive losses is that the former use exactly one positive and one negative pair per anchor and the positive pair of them is chosen from the same class and the negative pair is chosen from different classes. Contrastive learning generally uses just one positive pair for each anchor sample, selected using either co-occurrence~\cite{hjelm2018learning,henaff2020data} or data augmentation~\cite{chen2020simple}. The introduction of contrastive learning leads to a surge of interest in unsupervised visual representation learning~\cite{chen2020simple}. Wu et al.~\cite{wu2018unsupervised} maximize distinction between instances via a novel nonparametric softmax formulation and use a memory bank to store the instance class representation vector. For effective similarity measurement between samples in low-dimensional embedding space, other work explores the use of in batch samples for negative sampling instead of a memory bank~\cite{ye2019unsupervised,ji2019invariant}. Recently, researchers have attempted to relate the success of their methods to the maximization of mutual information between latent representations~\cite{bachman2019learning,henaff2020data}.

In probability theory and information theory, the mutual information of two random variables is a measure of their mutual dependence~\cite{mutualinformation}. It has important applications to contrastive learning~\cite{chen2020simple}. By maximizing mutual information between node and graph representations, some works, focusing on general graphs, have achieved state-of-the-art results in unsupervised node and graph classification tasks~\cite{velivckovic2018deep,sun2019infograph}. Maximizing mutual information between features extracted from multiple views of a shared context is analogous to human learning to represent observations generated by a shared cause driven by a desire to predict other related observations~\cite{bachman2019learning}. Aiming at a specific graph structure and task, we introduce a multi-view spatial-temporal graph contrastive representation learning method for view-invariant skeleton-based action recognition in this paper.

\begin{figure*}
	\centering
	\includegraphics[trim={0 0 0 0},clip,scale=0.6]{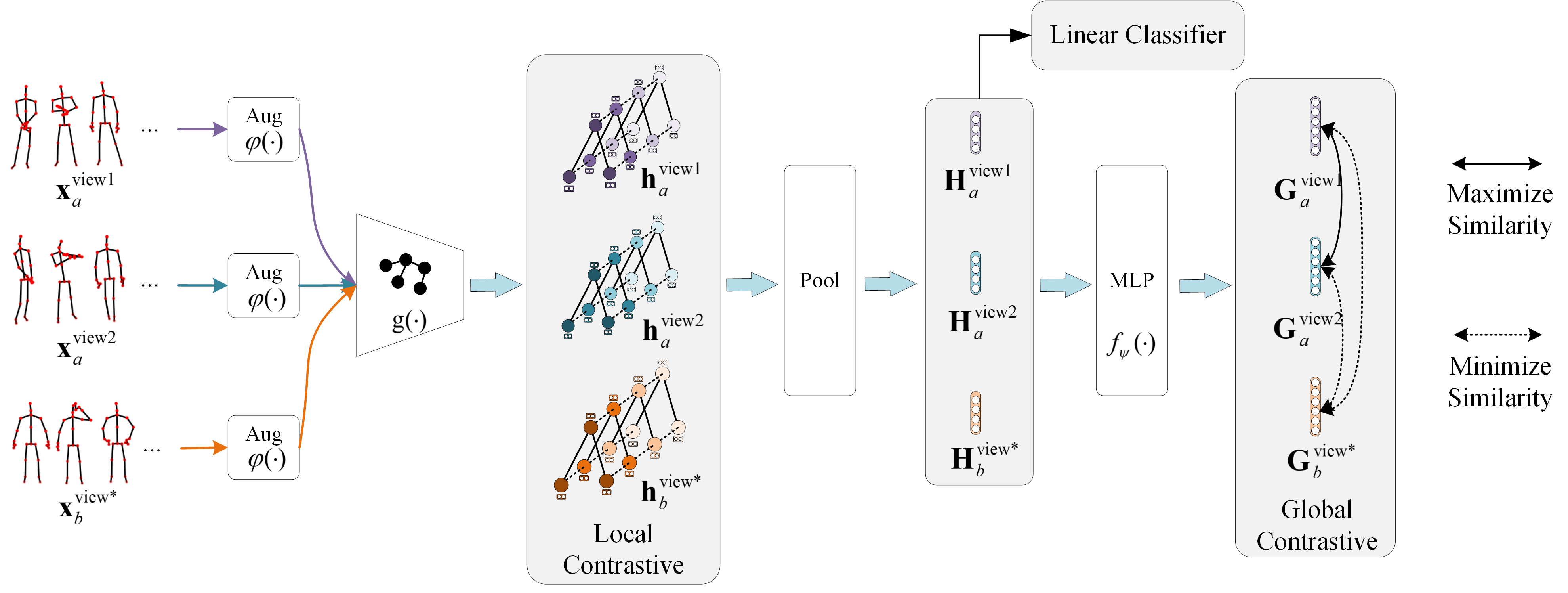}
	\caption{The overall pipeline of the proposed multi-view ST-Graph CRL for view-invariant skeleton-based action recognition. $\mathbf{x}_a^{\mathrm{view1}}$ and $\mathbf{x}_a^{\mathrm{view2}}$ are from any two views of the multi-view skeleton sequence $\mathbf{X}_a$. $\mathbf{x}_b^{\mathrm{view*}}$ is from any view of the multi-view skeleton sequence $\mathbf{X}_b$. This approach pulls together skeletons simultaneously captured for the same person from different views in embedding space, while pushing apart the others.}
	\label{method1}
	\vspace{0pt}
\end{figure*}
\section{Multi-View ST-Graph Contrastive Representation Learning}
\begin{figure*}
	\centering
	\includegraphics[trim={0 0 0 0},clip,scale=0.7]{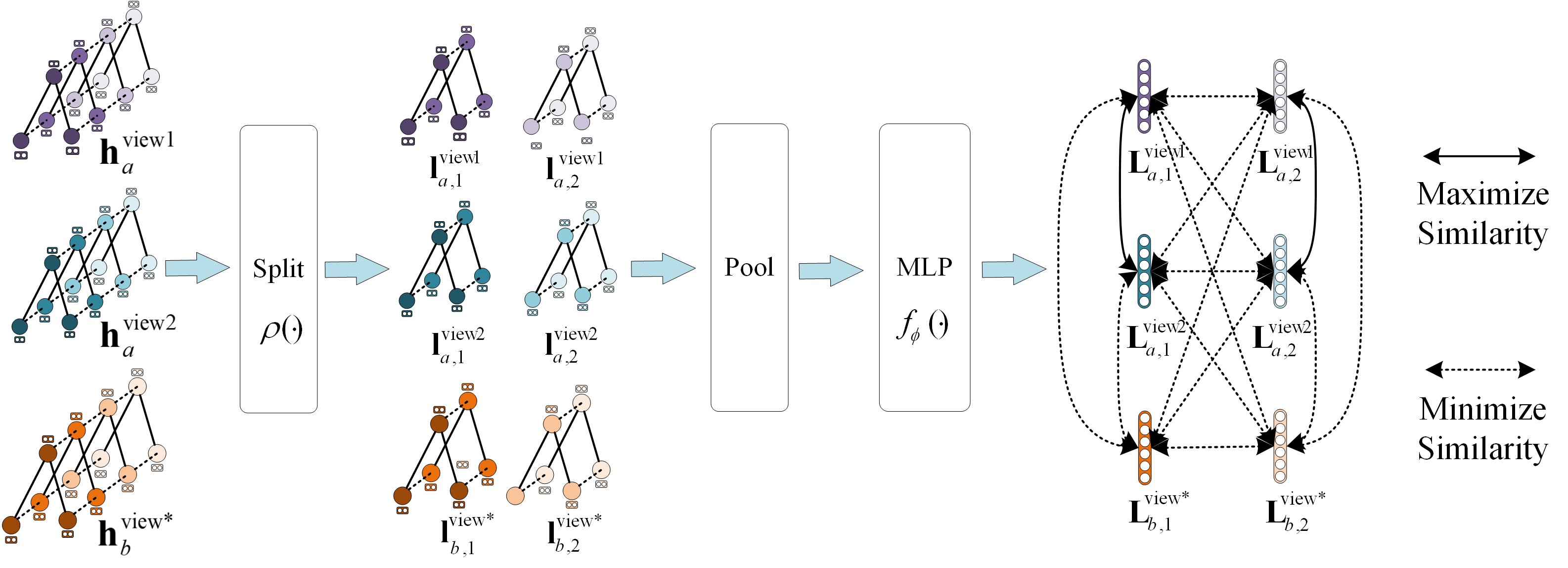}
	\caption{An more detail illustration of local contrastive in Figure~\ref{method1}. ST-Graph structural representations $\mathbf{h}_a^{\mathrm{view1}}$ and $\mathbf{h}_a^{\mathrm{view2}}$ are from any two views of the multi-view skeleton sequence $\mathbf{X}_a$ while $\mathbf{h}_b^{\mathrm{view*}}$ is from any view of the multi-view skeleton sequence $\mathbf{X}_b$. This loss aims to pull together skeleton sequence regions simultaneously captured for the same person from different views in embedding space, while pushing apart the others.}
	\label{method2}
	\vspace{0pt}
\end{figure*}

Inspired by recent contrastive learning algorithms, we propose an approach to learning view-invariant representations without any manual action labeling by maximizing the mutual information between skeleton sequences that are simultaneously taken for the same person but from different views, via a global-local contrastive loss in the latent space. The overall pipeline of the proposed approach is illustrated in Figure~\ref{method1}.
Specifically, a stochastic data augmentation module $\varphi(\cdot)$ that transforms any given data example randomly to encourage learning a more robust representation for the downstream task. Then, an ST-GCN structural encoder $g(\cdot)$ extracts representation vectors from augmented data examples. We maximize the representation agreement between samples simultaneously taken for the same person but from different views on a global level and a local level. On the global level, a small neural network projection head $f_\psi(\cdot)$ maps the representations to a latent space by applying a global contrastive loss. On the local level, as shown in Figure~\ref{method2}, an ST-Graph partitioning function $\rho(\cdot)$ splits the graph structural representation of the whole skeleton sequence into multi-local subgraphs, and then a projection head $f_\phi(\cdot)$ maps the representations to a latent space by applying a local contrastive loss. Moreover, to effectively combine global and local contrastive losses, we adjust their relative weights based on task uncertainty.

Before getting into the details of the approach, we state the main notations. Similar to previous studies~\cite{Cheng2020Skeleton, zhang2020semantics}, we organize skeleton sequence of an action sample as an undirected spatial-temporal graph $\mathbf{x} = (\mathbf{\mathcal{J},\mathcal{E}})$, where $\mathcal{J} = \{\mathbf{j}_{ti}\mid t=1,...,T; i=1,..., M\} $ denotes a set of vertices, corresponding to $T$ frames and $M$ body joints per frame, and $\mathcal{E}$ is the set of edges, indicating the connections between nodes. Then, we represent a multi-view skeleton sample as $\mathbf{X}=\{\mathbf{x}^v\}_{v=1}^{V}$, where $V$ represents the number of viewpoints, which could be as many as needed, and $v$ indicates the specific $v$-th viewpoint. For many multi-view skeleton samples,  we also use $\mathbf{x}_i^v$ to denote the $v$-th view of the $i$-th multi-view skeleton sample $\mathbf{X}_i$.

\subsection{Multi-view skeletal data augmentation.} Data augmentation aims to create novel and realistically rational data by applying a certain transformation to the original training data without affecting their semantic meanings. 
It has been demonstrated that contrastive learning usually needs stronger data augmentation than supervised learning~\cite{chen2020simple}. Meanwhile, for specific graphs, certain data augmentations might be more effective than the others~\cite{you2020graph}. 
Let an augmented skeleton sequence be $\hat{\mathbf{x}}_i^v = \varphi(\mathbf{x}_i^v)$, where $\varphi(\cdot)$ is the augmentation function. In this paper, we apply temporal subgraph as the data augmentation for multi-view CRL, with definitions as follows: it samples a segment from $\mathbf{x}_i^v$ along the temporal dimension. As the length of a skeleton sequence is fixed to 100 frames, we randomly sample 95 consecutive frames and then extend it to 100 frames by linear interpolation. This data augmentation increases the robustness of action recognition when the starting and ending frames of the action cannot be accurately determined and the skeleton sequences captured from different views do not have perfect temporal alignment.

\subsection{ST-GCN structural encoder.} The ultimate goal of ST-Graph CRL is to train a powerful skeleton sequence encoder $g(\cdot)$ to get view-invariant representation for skeleton-based action recognition without any manual action labeling. Specifically, to effectively model the co-occurrence relationships among joints in both spatial and temporal domains, we apply an ST-GCN structural encoder, which extracts representation $\mathbf{h}_i^v$ from the augmented skeleton sequence $\hat{\mathbf{x}}_i^v$. Specifically, it contains two parts: spatial graph convolution and temporal graph convolution.

For spatial graph convolution, the neighbor set of joints is defined as an adjacent matrix $\mathbf{A} \in \{0,1\}^{M \times M}$ according to $\mathcal{E}$, which is typically partitioned into 3 partitions: the centripetal group containing neighboring nodes that are closer to the skeleton center, the node itself and otherwise the centrifugal group. For individual skeleton, let $\mathbf{F} \in \mathbb{R} ^{M \times C}$ and $\mathbf{F}^{\prime} \in \mathbb{R} ^{M \times C^{\prime}}$ denote the input and output feature during the processing respectively, where $C$ and $C^{\prime}$ are the input and output feature dimensions. The graph convolution is computed as:
\begin{equation}
\mathbf{F}^{\prime} = \sum_{p\in\mathcal{P}} \bar{\mathbf{A}}_{p}\mathbf{F}\mathbf{W}_p,
\end{equation}
where $\mathcal{P} = \{\mathrm{root, centripetal, centrifugal}\}$ denotes the spatial partitions, $\bar{\mathbf{A}}_{p} = \mathbf{\Lambda}_p^{-\frac{1}{2}}\mathbf{A}_p \mathbf{\Lambda}_p^{-\frac{1}{2}}\in \mathbb{R} ^{M \times M}$ is the normalized adjacent matrix and $\mathbf{\Lambda}_p^{ij} = \sum_{j}(\mathbf{A}_p^{ij})+\alpha$, $\alpha$ is set to 0.001 to avoid empty rows. $\mathbf{W}_p\in\mathbb{R} ^{1\times1\times C \times C^{\prime}} $ is the weight of the $1\times1$ convolution for each partition group. For temporal dimension, we construct temporal graph by connecting identical joints in consecutive frames and use regular 1D convolution on the temporal dimension as the temporal graph convolution. 

The ST-GCN structural encoder comprises a series of dynamic spatial-temporal graph convolution blocks stacked one above the other.
In this form, there existed many specific models with subtle differences~\cite{yan2018spatial, shi2019two, Cheng2020Skeleton}. The proposed approach does not place any restriction on the ST-GCN structural encoder, as long as it maintains the feature of the spatial-temporal graph structure. In our implementation, we adopt the network recently proposed by~\cite{Cheng2020Skeleton} as the ST-GCN structural encoder.

\subsection{ST-Graph partitioning function.} 

As stated in Li et al.~\cite{li2018deeper}, the graph convolution operation can be considered Laplacian smoothing for node features over graph topology. The Laplacian smoothing computes the new node features as the weighted average of itself and its neighbors. It helps make nodes in the same cluster tend to learn similar representations. Nevertheless, it may also lead to the over-smoothing
problem and make nodes indistinguishable as the number of network layers increases. Meanwhile, it may concentrate more on node features and make the learned embeddings lack structural information. In short, ST-GGN can handle most simple cases but may ignore local details on a complicated graph. 

\begin{figure}
	\centering
	\includegraphics[trim={0 0 0 0},clip,scale=0.65]{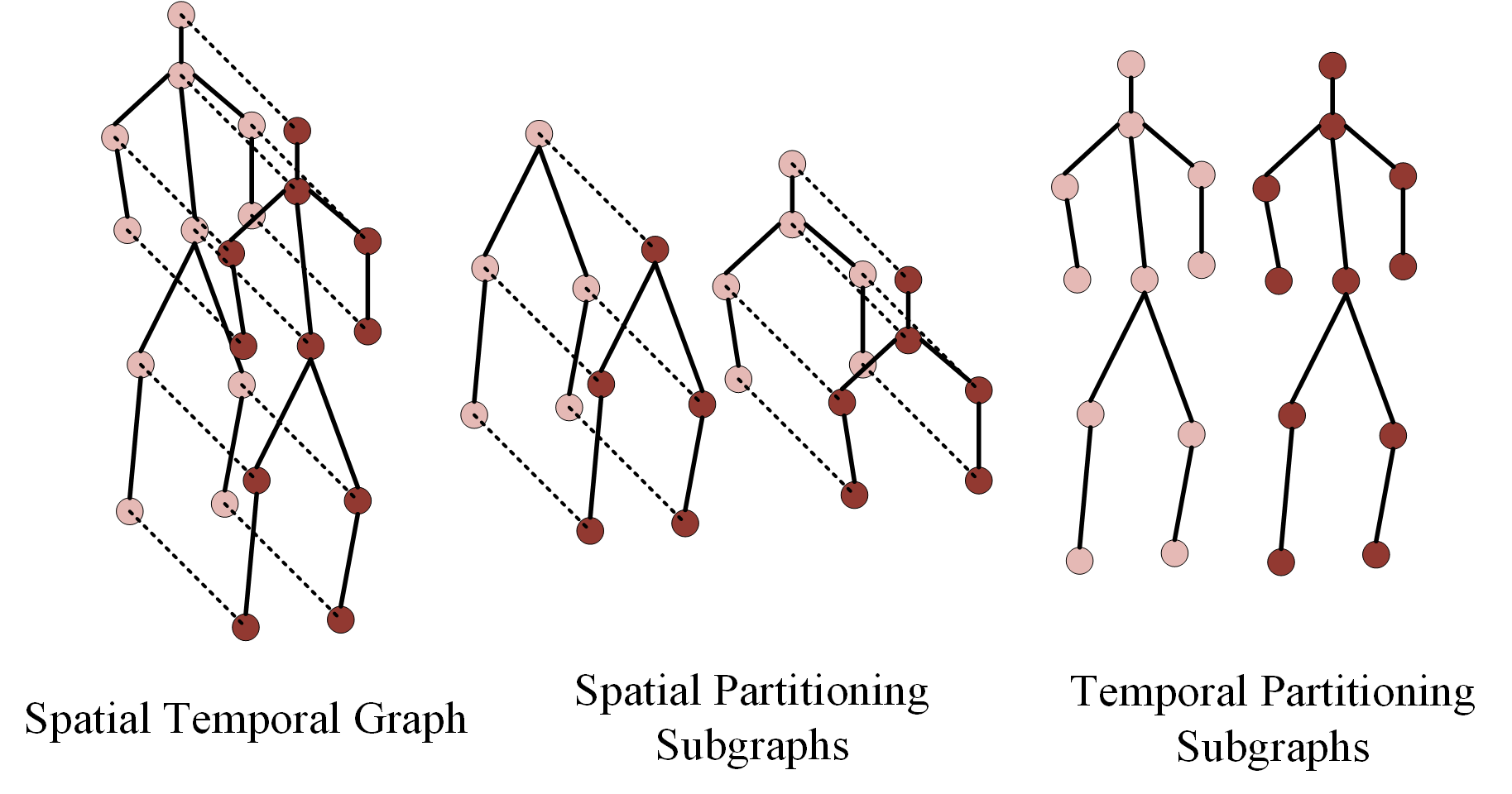}
	\caption{ST-Graph spatial or temporal partitioning strategies. The spatial-temporal feature graph are evenly partitioned along different dimensions by cutting edges.}
	\label{partition}
	\vspace{0pt}
\end{figure}

Given the above problems, we enhance the representation by giving more consideration to specific characteristics of local regions. Specifically, we include an ST-Graph partitioning function $\rho(\cdot)$ to split the feature of the whole skeleton sequence $\mathbf{h}_i^v$ into multi-local subgraphs $\mathbf{l}_{i,s}^v, s\in[1,...,S]$, where $S$ represents the number of generated subgraphs, $i$ and $s$ indicate sample index and subgraph index, respectively. The choice of partitioning strategies has a strong impact on not only the performance of recognition networks but also the design of the networks~\cite{fan2020application}. Several graph partitioning algorithms have already been developed and they are often either edge cut~\cite{andreev2006balanced}, which evenly partitions vertices and cuts edges, or vertex cut~\cite{bourse2014balanced}, which evenly partitions edges by replicating vertices. There have also been hybrid algorithms~\cite{li2019topox}, which cut both edges and vertices. In this paper, we adopt two simple rule-based edge cut style partitioning strategies to segment the skeleton spatial-temporal feature graph. Specifically, vertices of the ST-Graph are evenly partitioned into $S$ segments along the spatial dimension or the temporal dimension by cutting edges,  as shown in figure~\ref{partition}. 
\subsection{Projection head.} 
Recent work by Chen et al.~\cite{chen2020simple} found that mapping features to another latent space before contrastive loss calculation can be more effective. In this way, the features before a nonlinear projection are the learned representations, where information loss of raw data induced by the contrastive loss can be relieved. Therefore, in this paper, the representations $\mathbf{h}_i^v$ and $\mathbf{l}_{i,m}^v$ are mapped to another latent space through an MLP with one hidden layer, respectively. We name this module as projection head and add it to global and local contrastive learning subnetworks. Meanwhile, a global pooling is performed on $\mathbf{h}_i^v$ and $\mathbf{l}_{i,m}^v$ to get a fixed dimension feature vector for each ST-Graph to aggregate the node features before the projection head. Formally, the process is defined as:
\begin{equation}
\resizebox{0.9\linewidth}{!}{$
	\displaystyle
	\begin{split}
	\mathbf{G}_i^v = f_\psi(\mathrm{pool}(\mathbf{h}_i^v)) = \mathbf{W}^{(\psi, 2)}\sigma(\mathbf{W}^{(\psi, 1)}\mathrm{pool}(\mathbf{h}_i^v)), \\
	\mathbf{L}_{i,m}^v = f_\phi(\mathrm{pool}(\mathbf{l}_{i,m}^v)) = \mathbf{W}^{(\phi, 2)}\sigma(\mathbf{W}^{(\phi, 1)}\mathrm{pool}(\mathbf{l}_{i,m}^v)),
	\end{split}$}
\end{equation}
where $f_\psi(\cdot)$ and $f_\phi(\cdot)$ represent global and local projection heads. $\mathbf{G}_i^v$ and $\mathbf{L}_{i,m}^v$ are the global and local representations in another latent space. $\mathrm{pool}(\cdot)$ is a global pooling function. $\sigma$ is a ReLU nonlinearity and $\mathbf{W}$'s are learned weights of MLP. Note that the output of $\mathrm{pool}(\mathbf{h}_i^v)$ is named as $\mathbf{H}_i^v$, which is the representation we learned for the skeleton sequence based on ST-Graph CRL. 

\subsection{Global-Local contrastive learning.} 
A global representation can well capture the common knowledge of action patterns among all the regions in the skeleton sequence and hence possesses the nice merit in terms of model generalization while a local representation targets personalization of individual regions. As mentioned above, we propose several ST-Graph partitioning strategies to segment the graph into multiple local subgraphs. In this section, a global-local contrastive learning loss is proposed to effectively model the multi-scale co-occurrence relationship between spatial and temporal domains in the ST-Graph. For this, we define different positive pairs in global and local scenarios and maximize the consistency between the positive pairs compared with corresponding negative pairs using global and local contrastive loss functions. Meanwhile, the two contrastive loss functions are combined with task uncertainty in order to balance the trade-off between generalization and personalization of representation.

\textbf{Global contrastive loss.} Given two global representations $\mathbf{G}_a^{v1}$ and  $\mathbf{G}_b^{v2}$, we specify that they form a positive pair if $a$ is equal to $b$, else they form a negative pair. It means multiple skeleton sequences, if simultaneously taken for the same person from different views, will be pulled together in embedding space, otherwise will be pulled apart, which is shown in Figure~\ref{method1}. Therefore, not only skeleton representations can be effectively learned without any action label information, but also their view-invariant property of them can be enhanced during multi-view contrastive learning. To achieve this, we adopt the normalized temperature-scaled cross-entropy loss~\cite{chen2020simple}. Specifically, we randomly sample a minibatch of $N$ examples and define the contrastive prediction task on pairs of skeleton sequences. Note that each example consists of $V$ skeleton sequences collected from $V$ different views, resulting in $VN$ data points. Given $V$ positive pairs in an example, we treat the other $V(N-1)$ data points within a minibatch as negative examples. Let $\mathbf{u}$ and $\mathbf{v}$ denote representations of two data points. To measure similarity, we define $\mathrm{sim}(\mathbf{u}, \mathbf{v}) = \mathbf{u}^\top \mathbf{v} / \left\|\mathbf{u}\right\| \left\|\mathbf{v}\right\|$ that denotes the dot product between $\ell_2$ normalized $\mathbf{u}$ and $\mathbf{v}$. Then, the global loss function for positive pairs of example $i$ is defined as
\begin{equation}
\begin{aligned}
\resizebox{.91\linewidth}{!}{$
	\displaystyle
	\begin{split}
 & \ell_{i}^{\mathrm{global}}=-\log\\ &	\frac{\sum_{v1,v2=1}^{V}  \mathbbm{1}{[v1 \neq v2]}\exp \left(\operatorname{sim}\left(\mathbf{G}_{i}^{v1}, \mathbf{G}_{i}^{v2}\right) / \tau\right)}{\sum_{k=1}^{N}\sum_{v1,v2=1}^{V} \mathbbm{1}{\begin{tiny}
		\begin{bmatrix}
		k \neq i \\
		v1 \neq v2
		\end{bmatrix}
		\end{tiny}} \exp \left(\operatorname{sim}\left(\mathbf{G}_{i}^{v1}, \mathbf{G}_{k}^{v2}\right) / \tau\right)},
	\end{split}
$}
\end{aligned}
\end{equation}
where $\mathbbm{1}{[v1 \neq v2]} \in \{0, 1\}$ is an indicator function evaluating to 1 if $v1 \neq v2$, $\mathbbm{1}{\begin{tiny}
	\begin{bmatrix}
	k \neq i \\
	v1 \neq v2
	\end{bmatrix}
	\end{tiny}}$ is also an indicator function evaluating to 1 if $k \neq i$ and $v1 \neq v2$ are simultaneously satisfied, otherwise evaluating to 0. $\tau$ denotes a temperature parameter. For a minibatch, the global contrastive loss $\mathcal{L}_{\mathrm{global}}$ is computed across all examples,
\begin{equation}
\mathcal{L}_{\mathrm{global}} = \frac{1}{VN}\sum_{m=1}^{N}\ell_{m}^{\mathrm{global}},
\end{equation}
where $N$ is the batchsize.

\textbf{Local contrastive loss.} Local contrastive loss is calculated among the local representations, as illustrated in Figure~\ref{method2}. Given two local representations $\mathbf{L}_{a,s1}^{v1}$ and  $\mathbf{L}_{b,s2}^{v2}$, we specify that they form a positive pair if both $a=b$ and $s1=s2$ are satisfied, else they form a negative pair. From the composition of the positive and negative pairs, the contrastive loss achieves the same effect as the global one in the local scale when subgraph indices are consistent for all pairs. Besides, it can also handle the over-smoothing and the structural information lacking problems by contrasting among local regions in a sequence when sample indices are consistent for all pairs. The definition of local contrastive loss is the same as the global one. But because of the extra subgraph dimension, there are $VS$ positive pairs and $VS(N-1)$ negative pairs in a sample. Formally, the local contrastive loss function for positive pairs of example $i$ is defined as
\begin{equation}
\begin{aligned}
\resizebox{.91\linewidth}{!}{$
	\displaystyle
	\begin{split}
&\ell_{i}^{\mathrm{local}}=-\log\\&  \frac{\sum_{s=1}^{S}\sum_{v1,v2=1}^{V} \mathbbm{1}{[v1 \neq v2]} \exp \left(\operatorname{sim}\left(\mathbf{L}_{i,s}^{v1}, \mathbf{L}_{i,s}^{v2}\right) / \tau\right)}{\sum_{k=1}^{N}\sum_{s1,s2=1}^{S}\sum_{v1,v2=1}^{V}
	\mathbbm{1}{\begin{tiny}
		\begin{bmatrix}
		k \neq i \\
		s1 \neq s2\\
		v1 \neq v2
		\end{bmatrix}
		\end{tiny}} \exp \left(\operatorname{sim}\left(\mathbf{L}_{i,s1}^{v1}, \mathbf{L}_{k,s2}^{v2}\right) / \tau\right)},
	\end{split}
$}
\end{aligned}
\end{equation}
where $\mathbbm{1}{\begin{tiny}
	\begin{bmatrix}
	k \neq i \\
	s1 \neq s2\\
	v1 \neq v2
	\end{bmatrix}
	\end{tiny}}$ is an indicator function that needs the three inequalities are simultaneously true. $S$ is the number of split subgraphs. For a minibatch, the local contrastive loss $\mathcal{L}_{\mathrm{local}}$ also needs to be computed across all examples,
\begin{equation}
\mathcal{L}_{\mathrm{local}} = \frac{1}{SVN}\sum_{m=1}^{N}\ell_{m}^{\mathrm{local}}.
\end{equation}
Corresponding to spatially or temporally partitioning the ST-Graph into multiply subgraphs, the notation of local contrastive loss is $\mathcal{L}_{\mathrm{spalocal}}$ or $\mathcal{L}_{\mathrm{temlocal}}$. 

\textbf{Global-Local contrastive loss.} Global-Local contrastive loss is concerned about jointly optimizing the related global and local contrastive loss functions. In this paper, the popular approach of using a linear combination of them as a total loss function is abandoned. Because manually tuning their weight hyper-parameters is expensive and intractable. Instead, following the work of~\cite{wang2020m2grl}, we adjust each loss's relative weight in the total loss function by deriving a multi-task loss function based on maximizing the Gaussian likelihood with task-dependent uncertainty during model training. We define the global-local contrastive loss $\mathcal{L}$ as follows:
\begin{equation}
\mathcal{L} = \frac{1}{\sigma_1^2}\mathcal{L}_{\mathrm{global}} + \frac{1}{\sigma_2^2}\mathcal{L}_{\mathrm{local}}+
\log(\sigma_1^2) + \log(\sigma_2^2), 
\end{equation}
where $\sigma_1$ and $\sigma_1$ associate with the task uncertainty and can be interpreted as the relative weights of respective loss terms. $\log(\sigma_1^2)$  and $\log(\sigma_2^2)$ serve as regularizers to avoid over-fitting. All network parameters and the uncertainty task weights are trainable and optimized by gradient back propagation.

The proposed multi-view ST-Graph CRL is summarized as Algorithm~\ref{alg}.
\begin{algorithm}
	\caption{Multi-view spatial-temporal graph contrastive representation learning algorithm}
	\label{alg}
	\textbf{Input}: Augmentation $\varphi(\cdot)$, global pooling $\mathrm{pool}(\cdot)$, ST-Graph partitioning function $\rho(\cdot)$, ST-GCN structural encoder $g(\cdot)$, global and local projection heads $f_\psi(\cdot)$ and $f_\phi(\cdot)$, training multi-view skeleton sequences $\{\mathbf{X}_i = \{\mathbf{x}_i^v\}_{v=1}^{V} \}_{i=1}^{N}$, global contrastive loss $\mathcal{L}_{\mathrm{global}}$, local contrastive loss $\mathcal{L}_{\mathrm{local}}$,  similarity measurement function $sim(\cdot)$.\\
	\textbf{Parameters}: Learnable relative weight parameters for global and local contrastive loss: $\sigma_1$ and $\sigma_2$; number of views $V$; number of split subgraphs $S$; number of samples in one batch $K$; temperature parameter $\tau$.
	\begin{algorithmic}[1] 
		\WHILE{sampled batch~$\{\{\mathbf{x}_i^v\}_{v=1}^{V}\}_{i=1}^{K}$}
		\WHILE{$i=1$ to $K$}
		\WHILE{$v=1$ to $V$}
		\STATE  $\mathbf{h}_i^v = g(\varphi(\mathbf{x}_i^v))$
		\STATE $\mathbf{G}_i^v = f_\psi(\mathrm{pool}(\mathbf{h}_i^v))$
		\STATE $\{\mathbf{l}_{i,s}^v\}_{s=1}^S =  \rho(\mathbf{h}_i^v)$
		\WHILE{$s=1$ to $S$}
		\STATE $\mathbf{L}_{i,s}^v = f_\phi(\mathrm{pool}(\mathbf{l}_{i,s}^v))$
		\ENDWHILE
		\ENDWHILE
		\ENDWHILE 
		\WHILE{$i=1$ to $K$}
		\STATE
		$\ell_{i}^{\mathrm{global}}=-\log$
		$\frac{\sum_{v1,v2=1}^{V}  \mathbbm{1}{[v1 \neq v2]}\exp \left(\operatorname{sim}\left(\mathbf{G}_{i}^{v1}, \mathbf{G}_{i}^{v2}\right) / \tau\right)}{\sum_{k=1}^{K}\sum_{v1,v2=1}^{V} \mathbbm{1}{\begin{tiny}
				\begin{bmatrix}
				k \neq i \\
				v1 \neq v2
				\end{bmatrix}
				\end{tiny}} \exp \left(\operatorname{sim}\left(\mathbf{G}_{i}^{v1}, \mathbf{G}_{k}^{v2}\right) / \tau\right)}$	
		\STATE
		$\ell_{i}^{\mathrm{local}}=-\log$ 
		$\frac{\sum_{s=1}^{S}\sum_{v1,v2=1}^{V} \mathbbm{1}{[v1 \neq v2]} \exp \left(\operatorname{sim}\left(\mathbf{L}_{i,s}^{v1}, \mathbf{L}_{i,s}^{v2}\right) / \tau\right)}{\sum_{k=1}^{K}\sum_{s1,s2=1}^{S}\sum_{v1,v2=1}^{V}
			\mathbbm{1}{\begin{tiny}
				\begin{bmatrix}
				k \neq i \\
				s1 \neq s2\\
				v1 \neq v2
				\end{bmatrix}
				\end{tiny}} \exp \left(\operatorname{sim}\left(\mathbf{L}_{i,s1}^{v1}, \mathbf{L}_{k,s2}^{v2}\right) / \tau\right)}$
		\ENDWHILE
		\STATE $\mathcal{L}_{\mathrm{global}} = \frac{1}{VK}\sum_{m=1}^{K}\ell_{m}^{\mathrm{global}}$
		\STATE $\mathcal{L}_{\mathrm{local}} = \frac{1}{SVK}\sum_{m=1}^{K}\ell_{m}^{\mathrm{local}}$
		\STATE $\mathcal{L} = \frac{1}{\sigma_1^2}\mathcal{L}_{\mathrm{global}} + \frac{1}{\sigma_2^2}\mathcal{L}_{\mathrm{local}}+
		\log(\sigma_1^2) + \log(\sigma_2^2)$
		\STATE update networks $g(\cdot)$, $f_\psi(\cdot)$, $f_\phi(\cdot)$, $\sigma_1$ and $\sigma_2$ to minimize $\mathcal{L}$
		\ENDWHILE
		\STATE \textbf{return} encoder model $g(\cdot)$, and throw away projection heads $f_\psi(\cdot)$ and $f_\phi(\cdot)$
	\end{algorithmic}
\end{algorithm}

\section{Experiment}


\subsection{Dataset}
We evaluate the proposed method on two public available multi-view action recognition benchmarks: NTU RGB+D~\cite{shahroudy2016ntu} and PKUMMD~\cite{Liu2017PKU}. We briefly describe them below.

\textbf{NTU RGB+D (NTU).} NTU is a large-scale multi-modal action recognition dataset. It is composed of 56,880 samples over 60 classes captured from 40 distinct subjects and three Kinect cameras. Each action in the samples involves one or two people. The dataset is very challenging due to the large intra-class and view variations. The original paper of the NTU recommends two benchmarks: 1) Cross-subject (CS): all samples from a selected group of subjects are used for training and the rest samples for testing. 2) Cross-view (CV): the training set contains samples that are captured
by cameras 2 and 3, and the testing set contains videos
that are captured by camera 1. We follow this convention and report performance on both benchmarks.

\textbf{PKUMMD.} PKUMMD is a new large-scale benchmark for continuous multi-modality 3D human action understanding and covers a wide range of complex human activities with well-annotated information. It contains almost 20,000 action instances in 51 action categories, performed by 66 subjects in three different view Kinect sensors. PKUMMD consists of two subsets: PKUMMD-I is an easier subset for action recognition, while PKUMMD-II is more challenging with more skeleton noise caused by large view variation. We conduct experiments under the cross-subject protocol on the two subsets.

\subsection{Implementation Details}
\subsubsection{Pre-training without any action label information}
In ST-Graph CRL, an ST-GCN structural encoder $g(\cdot)$, a global projection head $f_\phi(\cdot)$ and a local projection head $f_\psi(\cdot)$ are pre-trained using multi-view skeleton sequences without any action label information. We use SGD with Nesterov momentum 0.9 to pre-train them for 40 epochs. The learning rate is set to 0.1 and divided by 10 at epoch 20, 30, and 35. The batch size is set to 16 for all experiments. The sequence length $T$ is set to 100. The temperature parameter for global-local contrastive loss is set to 0.07. The number of subgraph $S$ is set to 5. $V$ is set to 2, which means each sample includes two skeleton sequences, simultaneously taken from different views. 

\subsubsection{Evaluation Protocol}
To validate the effectiveness of the proposed representation learning method, we follow the linear evaluation protocol~\cite{wang2020m2grl,chen2020simple}, which is commonly used to evaluate unsupervised learning methods. In this way, a linear classifier attached to the frozen encoder model $g(\cdot)$ is trained with the annotated dataset. We report Top-1 accuracy on the testing set as a quantitative evaluation indicator. The classifier is trained for 45 epochs, with the learning rate divided by 10 at epoch 25, 35, and 40. The other settings remain the same as the pre-training.

\subsection{Comparison Experiments}
To quantitatively evaluate the performance, Table~\ref{state-of-the-art} and Table~\ref{crossdataset} list the linear evaluation results of ST-Graph CRL and other state-of-the-art unsupervised methods on PKUMMD and NTU benchmarks. The model which only trains the linear classifier and freezes the randomly initialized encoder is denoted as ST-Graph Rand. We regard this model as one of our baselines. The models implementing ST-Graph contrastive learning in single-view and multi-view scenarios are denoted as ST-Graph CRL SV and ST-Graph CRL MV, respectively. In the single view version, we maximize the mutual information between skeleton ST-Graph representations of one augmented instance and another augmented instance of an identical skeleton sequence, to learn inherent action patterns of different skeleton transformations. For the evaluation of P\&C FW on the action recognition task, we reproduce the coder of P\&C FW with a linear evaluation protocol. The temporal subgraph is the default data augmentation method we adopt in these experiments. 

\begin{table*}
	\caption{Comparison of action recognition performance of the proposed ST-Graph CRL and other state-of-the-art methods.}
	\centering
	\begin{tabular}{clrrrrr}
		\toprule
		Supervised &Models  &  PKUMMD-I & PKUMMD-II &  NTU (CS) & NTU (CV)\\
		\midrule
		Yes	& ST-Graph & 94.45  & 56.75 & 87.82 & 95.13 \\
		
		No &ST-Graph Rand         & 30.14 & 10.58  & 19.55 & 23.23    \\
		No&LongT GAN~\cite{zheng2018unsupervised}    & 67.70 & 25.95  & 52.14 & -    \\
		No&P\&C FW~\cite{su2020predict}              & 67.62 & 35.90  & 32.50 & 35.67  \\
		No&$\mathrm{M}^2$SL~\cite{lin2020ms2l}       & 64.86 & 27.63  & 52.55 & -    \\
		No&CAE+~\cite{rao2021augmented}       & - & -  & 58.50 & 64.80    \\
		No&ST-Graph CRL SV (\textbf{Ours})         &  68.42 & 31.80  & 60.24 & 59.79    \\
		No&ST-Graph CRL MV(\textbf{Ours})            &  \textbf{83.62} & \textbf{39.89}  & \textbf{74.71} & \textbf{82.62}    \\
		\bottomrule
	\end{tabular}
	\label{state-of-the-art}
	\vspace{0pt}
\end{table*}
\subsubsection{Comparison with State-of-the-art}  In existing studies~\cite{su2020predict,lin2020ms2l}, the pre-training and evaluation are usually conducted on the same dataset. An overall summary of the results is given in Table~\ref{state-of-the-art}, where the proposed method has returned significantly improved performance in the unsupervised methods that do not use action labels for training. As we can see, ST-Graph CRL MV is far beyond the performance of random baseline and other state-of-the-art unsupervised methods and greatly reduces the gap to the models trained with action annotation. NTU(CV) is a suitable benchmark to evaluate the model's robustness to the viewpoint difference. Here, we can see that model’s Top-1 accuracy of ST-Graph CRL MV in NTU(CV) is 82.62\%, while ST-Graph Rand and P\&C FW are only 23.23\% and 35.67\%, respectively. Therefore, the multi-view contrastive learning significantly improved the view-invariant property of skeleton representation. Even in a single view scenario, under the truly unsupervised setting, the performances of ST-Graph CRL SV are quite outstanding, which performs better than almost all the baselines. It achieves high recognition accuracies of 60.21\% and 59.79\% on NTU(CS) and NTU(CV), respectively, which proves that our global-local contrastive learning of augmented skeletons of the same sample also works well. From the comparison of ST-Graph CRL SV and MV, we can see that substantial improvements are made in each benchmark. It proves that CRL between the multi-view skeletons brings in a giant performance leap for unsupervised skeleton-based action recognition.
\begin{table*}
	\caption{Performance of transfer learning setting and linear evaluation.}
	\centering
	\begin{tabular}{clrrr}
		\toprule
		Supervised & Models  & PKUMMD-I & PKUMMD-II\\
		\midrule
		Yes & ST-Graph & 90.56  & 55.01 \\
		No & P\&C FW~\cite{su2020predict}    & 63.31  & 23.61   \\
		No & ST-Graph CRL SV (\textbf{Ours}) & 76.29  & 39.83    \\
		No & ST-Graph CRL MV(\textbf{Ours})    & \textbf{82.21}  & \textbf{46.98}    \\
		\bottomrule
	\end{tabular}
	\label{crossdataset}
	\vspace{0pt}
\end{table*}

\subsubsection{Transfer learning performance.} To further evaluate whether the proposed ST-Graph CRL can gain knowledge of related tasks, we investigate the transfer learning performance of our model~\cite{lin2020ms2l}. As the representations learned from large-scale data are more generalizable, we regard the NTU as the source dataset and PKUMMD-I and PKUMMD-II as the target datasets. We conduct the pre-training on the source dataset and the evaluation of target datasets. Under this setting, the samples used for pre-training and linear evaluation are completely different in terms of viewpoints, action patterns, and so on, which is more in accordance with the practical scenarios. The results are summarized in Table~\ref{crossdataset}, from which ST-Graph CRL MV gets better results of 82.21\% for PKUMMD-I and 46.98\% for PKUMMD-II when models are pre-trained without action annotations. Apart from that, together with Table~\ref{state-of-the-art}, we can see that the accuracies of P\&C FW are reduced from 67.62\% and 35.90\% to 63.31\% and 23.61\%, respectively, while most ST-Graph CRLs are boosted from 68.42\%, 31.80\%, 83.62\%, and 39.89\% to 76.29\%, 39.83\%, 82.21\%, and 46.98\%, respectively, when the training and testing datasets are from consistent to inconsistent. Meanwhile, the performances of models pre-trained with action annotations also decrease in this transfer learning setting from 94.45\% and 56.75\% to 90.56\% and 55.01\%, respectively. One possible reason is that ST-Graph CRL can take advantage of a larger training set more effectively with less influence from the data distribution difference between different datasets. It can be concluded that the representations learned in ST-Graph CRL have a good generalization ability.

\subsection{Ablation Experiments}
For a specific ST-Graph structural encoder, the performance of ST-Graph CRL is mainly determined by the following four components: multi-view skeleton contrastive mechanism, data augmentation, projection head, and global-local contrastive loss. From the results of ST-Graph CRL SV and ST-Graph CRL MV in Table~\ref{state-of-the-art} and Table~\ref{crossdataset}, the performance of the multi-view skeleton contrastive mechanism is shown to be impressive in all cases. To further assess the other factors, we conduct several ablation experiments on NTU with a linear evaluation protocol.

\begin{table}
	\caption{Analysis of global-local contrastive learning loss function.}
	\centering
	\begin{tabular}{lrr}
		\toprule
		Loss Function  & NTU (CS) & NTU (CV) \\
		\midrule
		$\mathcal{L}_{\mathrm{global}}$    & 69.69  & 73.70     \\
		$\mathcal{L}_{\mathrm{temlocal}}$  & 67.36  & 74.88     \\
		$\mathcal{L}_{\mathrm{spalocal}}$  & 61.04  & 65.89     \\
		$\mathcal{L}_{\mathrm{global}} + \mathcal{L}_{\mathrm{temlocal}}$   & \textbf{74.71}  & \textbf{82.62}     \\
		$\mathcal{L}_{\mathrm{global}} + \mathcal{L}_{\mathrm{spalocal}}$   & 73.62  & 79.04     \\
		$\mathcal{L}_{\mathrm{global}}+ \mathcal{L}_{\mathrm{temlocal}} + \mathcal{L}_{\mathrm{spalocal}}$   & 74.21  & 81.54     \\
		\bottomrule
	\end{tabular}
	\label{globallocalloss}
	\vspace{0pt}
\end{table}
\begin{table}
	\caption{Performance by using different methods to combine the global and local contrastive learning losses.}
	\centering
	\begin{tabular}{llrr}
		\toprule
		Viewpoint  & Loss combination & NTU (CS) &NTU (CV)\\
		\midrule
		Single-view & Linear            & 54.12    & 56.23  \\
		& Task uncertainty  & 60.24    & 59.79  \\
		Multi-view  & Linear            & 71.06   & 78.68   \\
		& Task uncertainty  & \textbf{74.71}  & \textbf{82.62}   \\
		\bottomrule
	\end{tabular}
	\label{architecture}
	\vspace{0pt}
\end{table}
\subsubsection{The effect of global-local contrastive loss.} In this experiment, we evaluate different forms of the contrastive loss function. Experimental results are summarized in Table~\ref{globallocalloss}. Based on the results, we make the following observations. As the accuracy of $\mathcal{L}_{\mathrm{temlocal}}$ is higher than $\mathcal{L}_{\mathrm{spalocal}}$ by 6.32\%(CS) and 8.99\%(CV), the temporal splitting method is superior to the spatial splitting in this experiment for local contrastive loss. The impacts of the global and the local losses are different but complementary. Compared with using only one of them, the combined global-local loss function $\mathcal{L}_{\mathrm{global}} + \mathcal{L}_{\mathrm{temlocal}}$ leads to substantially better performance in two benchmarks, i.e., 74.71\%(CS) and 82.62\%(CV). 
In Table~\ref{globallocalloss}, it can be found that $\mathcal{L}_{\mathrm{spalocal}}$ only produces poor performance. We think the reason might be related to the multi-view action
datasets. Specifically, as most multi-view action datasets do not provide the corresponding relation between persons in different views, the spatial partitioning strategy is likely to lead to a phenomenon that the positive pairs of local parts are from different persons when an action is performed by two people. In this case, the effect of $\mathcal{L}_{\mathrm{spalocal}}$ is inconsistent with our expectation and one of its impacts, learning a fine-grained view of irrelevant representation in the spatial dimension, will fail, while others still work. However, when combined with $\mathcal{L}_{\mathrm{global}}$ and $\mathcal{L}_{\mathrm{temlocal}}$, the main role of $\mathcal{L}_{\mathrm{spalocal}}$ is reflected in learning a fine-grained view irrelevant representation in the spatial dimension. This is why its accuracy goes down. Therefore, the default form of global-local contrastive loss consists of $\mathcal{L}_{\mathrm{global}}$ and $\mathcal{L}_{\mathrm{temlocal}}$, combined with task uncertainty in this paper. We also compare linear and task uncertainty based methods to combine the global and local contrastive losses. Note that all the weight parameters are uniformly set to 1 in the linear combination method. Results are shown in Table~\ref{architecture}, from which we can see that the task uncertainty based combination method outperforms the linear combination methods in both single-view and multi-view scenarios. 

\subsubsection{Analysis of the projection head.} We study the importance of including a projection head, i.e. $f_\psi(\cdot)$ and $f_\phi(\cdot)$. Table~\ref{projectionhead} shows the linear evaluation results using two different architectures for the head: identity mapping and the nonlinear projection with one additional hidden layer. We can observe that a nonlinear projection head, regardless of its output representation dimension, performs better than identity mapping in terms of recognition accuracy. Therefore, it can be concluded that the hidden layer before the projection head is a better representation than the layer after.

\begin{table*}
	\caption{Linear evaluation of representations with different projection heads and various dimensions of output. The representation, before projection, is 256-dimensional here.}
	\centering
	\begin{tabular}{lcccccc}
		\toprule
		Projection head  & Identity Mapping & \multicolumn{5}{c}{Nonlinear projection}\\
		\midrule
		Output dimension & 256 & 32 & 64 & 128 & 256 & 512\\
		\midrule
		Accuracy & 66.21 & 74.29 & 74.31 & 74.39 & 74.38 & \textbf{74.71}  \\
		\bottomrule
	\end{tabular}
	\label{projectionhead}
	\vspace{0pt}
\end{table*}

\subsubsection{The effects of data augmentation.} 
Apart from the temporal subgraph, we also explore other four popular skeleton data augmentations in experiments including node dropping, node perturbation, view rotation, and skeleton shearing, with definitions as follows:

Node dropping. It randomly discards body joints in the input skeleton sequence $\mathbf{x}_i^v$. Specifically, with a $50\%$ chance, we randomly drop $10\%$ of nodes, where the corresponding joint coordinates are set to zero. It is a common phenomenon that a subset of joints, e.g., those occluded ones, cannot be detected. The augmentation of node dropping enables the crucial action patterns can still be learned from a subset of joints.

Node perturbation. The coordinates of joints are perturbed using a normal Gaussian distribution. The mean of the distribution is set to 0 while the standard deviation is set to 0.05. The detected joint locations, even for those without occlusion, always contain errors due to sensor and estimation accuracies in practice. The augmentation of node perturbation enables the action recognition to be robust to such errors. 

View rotation. It randomly rotates the joint coordinates in a skeleton sequence along three axes in terms of a rotation matrix. Specifically, we randomly select three degrees $\alpha, \beta, \gamma$, all uniformly in the range of $[-17^\circ, 17^\circ]$ for each sequence. Three basic rotation matrices with rotation angles about X, Y, Z axis are given as follows:

\begin{equation}
\begin{aligned}
\resizebox{.5\linewidth}{!}{$
	\displaystyle
	\begin{split}
& \mathbf{R}_{\mathrm{X}}(\alpha) = \begin{bmatrix}
1 & 0 & 0 \\
0 & \mathrm{cos}\alpha & \mathrm{sin}\alpha \\
0 & -\mathrm{sin}\alpha & \mathrm{cos}\alpha
\end{bmatrix},\\&
\mathbf{R}_{\mathrm{Y}}(\beta) = \begin{bmatrix}
\mathrm{cos}\beta & 0 & -\mathrm{sin}\beta \\
0 &1 &0 \\
\mathrm{sin}\beta & 0 & \mathrm{cos}\beta
\end{bmatrix},\\&
\mathbf{R}_{\mathrm{Z}}(\gamma) = \begin{bmatrix}
\mathrm{cos}\gamma & \mathrm{sin}\gamma & 0 \\
-\mathrm{sin}\gamma & \mathrm{cos}\gamma & 0 \\
0 & 0 & 1
\end{bmatrix}.
	\end{split}
$}
\end{aligned}
\end{equation}
Based on these three basic rotation matrices, the final rotation matrix is 
\begin{equation}
\mathbf{R} = \mathbf{R}_{\mathrm{X}}(\alpha)\mathbf{R}_{\mathrm{Y}}(\beta)\mathbf{R}_{\mathrm{Z}}(\gamma).
\end{equation}
We apply the rotation matrix $\mathbf{R}$ to the original coordinates of the skeleton sequence and get the transformed coordinates. It simulates the view changes of the camera. This augmentation enables the action recognition to be robust to the camera view changes.

Skeleton shearing. It slants the shape of a skeleton at a random angle. The shearing factors are drawn from a uniform distribution in $[0.01, 0.1]$. The transformation matrix can be written as
\begin{equation}
\mathbf{S} = \begin{bmatrix}
1 & s_\mathrm{X}^\mathrm{Y} & s_\mathrm{X}^\mathrm{Z} \\
s_\mathrm{Y}^\mathrm{X} & 1 & s_\mathrm{Y}^\mathrm{Z} \\
s_\mathrm{Z}^\mathrm{X} & s_\mathrm{Z}^\mathrm{Y} & 1
\end{bmatrix},
\end{equation}
where $s_\mathrm{X}^\mathrm{Y}$,  $s_\mathrm{X}^\mathrm{Z}$, $s_\mathrm{Y}^\mathrm{X}$, $s_\mathrm{Y}^\mathrm{Z}$, $s_\mathrm{Z}^\mathrm{X}$, $s_\mathrm{Z}^\mathrm{Y}$ are shearing factors.
All joint coordinates of the original skeleton sequence are transformed with the shearing matrix $\mathbf{S}$. The augmentation of skeleton shearing further increases the robustness of action recognition to more nonrigid transformations of the skeleton sequence.

\begin{table}
	\caption{Performance of multi-view ST-Graph~CRL using different augmentation strategies.}
	\centering
	\begin{tabular}{lrr}
		\toprule
		Augmentation  & NTU (CS) & NTU (CV) \\
		\midrule
		Original            & 69.85  & 78.35     \\
		Node Dropping       & 69.05  & 77.37      \\
		Node perturbation   & 66.26  & 74.99     \\
		View rotation       & 70.37  & 77.99    \\
		Shear               & 69.56  & 77.55     \\
		Temporal subgraph           & \textbf{74.71}  & \textbf{82.62}     \\
		\bottomrule
	\end{tabular}
	\label{augmentation}
	\vspace{0pt}
\end{table}

We denote the model without any data augmentation as the original. The results are shown in Table~\ref{augmentation}. Much to our surprise, compared with directly using the original sequence, only the temporal subgraph strategy to ST-Graph CRL can significantly improve the accuracy by 5.26\%(CS) and 4.27\%(CV). Two possible reasons are: 1) defining precise frame-level starting and ending time for action is almost impossible, and 2) it is hard to achieve a strict temporal alignment of skeleton sequences captured by multiple cameras. Applying inconsistent temporal subgraphs for different views can improve the robustness of these unavoidable problems without breaking their original relationships. The counterproductive of other data augmentations maybe because the original correspondences in spatial structure among skeletons, which are simultaneously taken from different views, are destroyed after these random transformations. For example, compared with other augmentations, node perturbation, which changes the values of joints with a normal Gaussian distribution, is most damaging to spatial structure correspondences and leads to the sharpest performance drop of 3.59\%(CS) and 3.36\%(CV). Thus, the temporal subgraph is the default data augmentation we adopt in this work.


\subsection{Visualization of Skeleton Representation.} 
\begin{figure*}
	\centering
	\subfigure[P\&C FW~\cite{su2020predict}]{\includegraphics[trim={80 80 30 80},clip,scale=0.3]{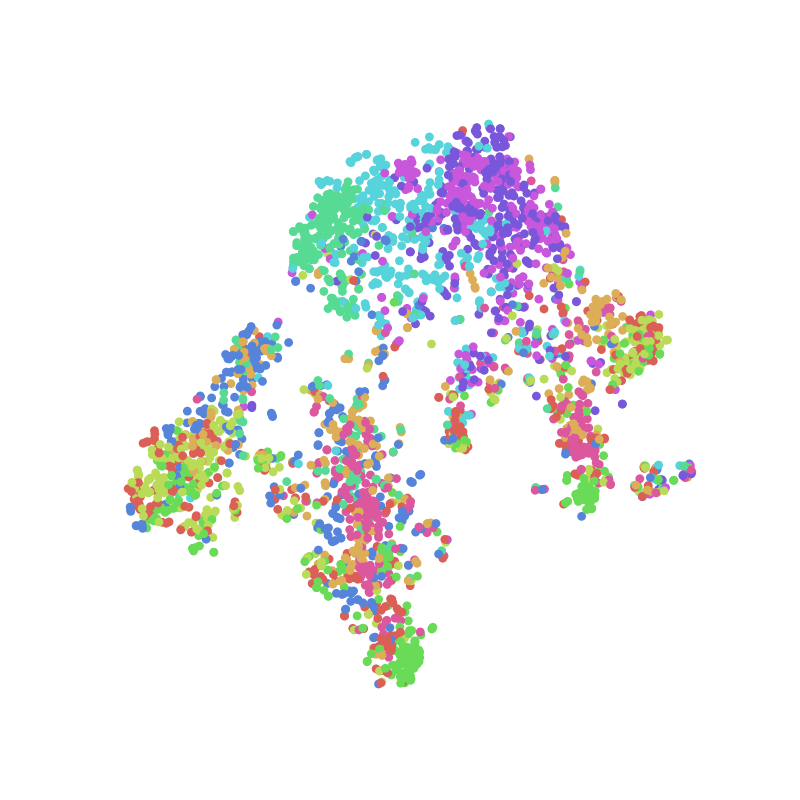}}
	\subfigure[ST-Graph CRL SV]{\includegraphics[trim={80 80 30 80},clip,scale=0.3]{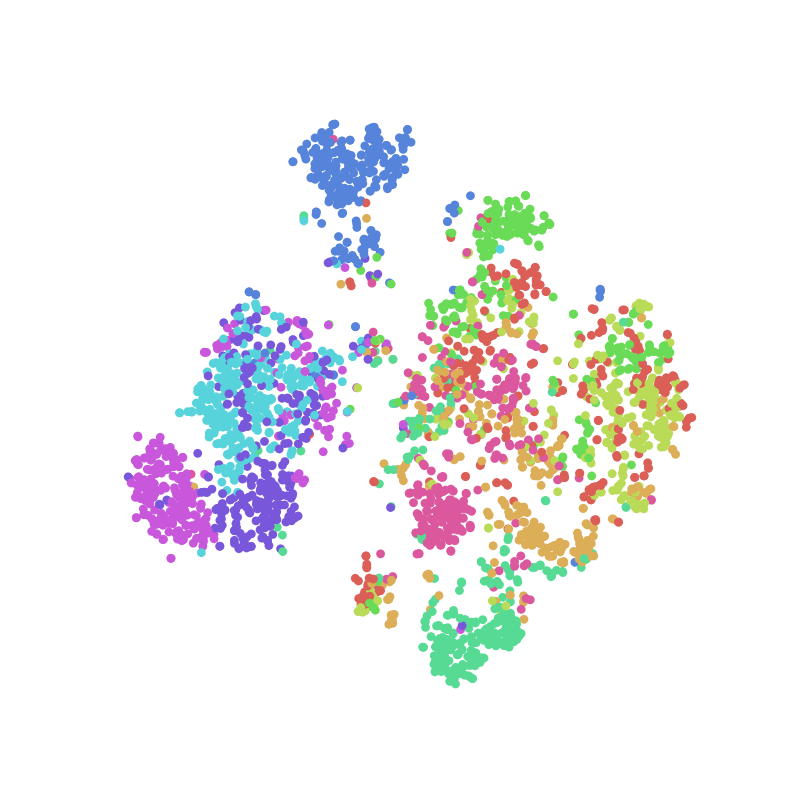}}
	\subfigure[ST-Graph CRL MV]{\includegraphics[trim={30 80 80 80},clip,scale=0.3]{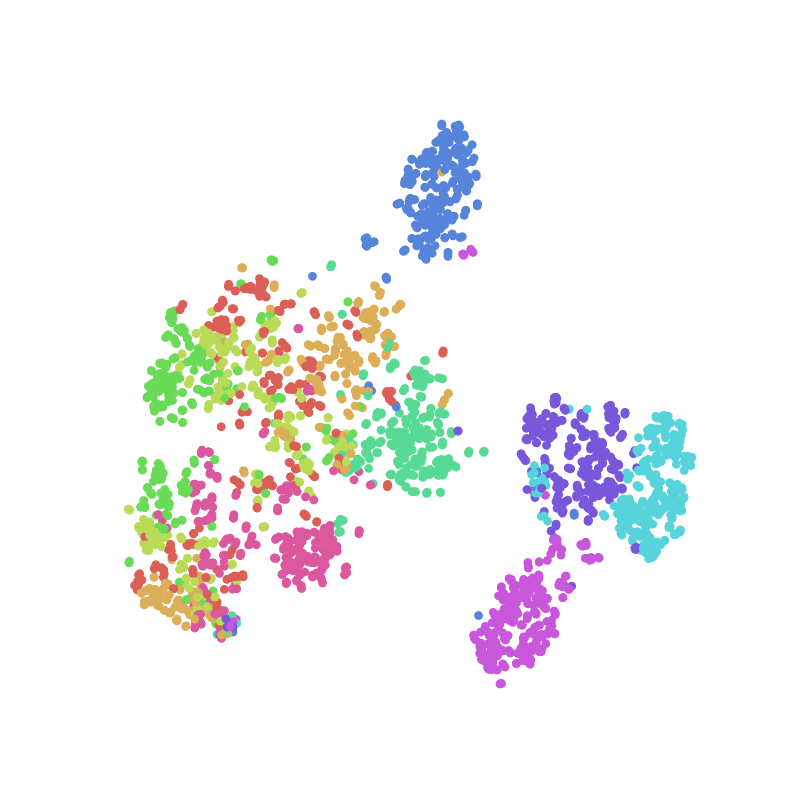}}
	\vspace{-10pt}
	\caption{TSNE-embedding visualizations of the learned representations from 10 classes randomly selected in NTU(CS) testing set.}
	\label{tsne}
	\vspace{0pt}
\end{figure*}

The superior performance of ST-Graph CRL over the existing methods is largely due to the use of the multi-view skeleton contrastive mechanism. Hence apart from the quantitative evaluation, we also visualize the feature changes by using this mechanism. We randomly select ten classes in the NTU testing set and visualize the TSNE-embeddings of the features obtained from P\&C FW~\cite{su2020predict}, ST-Graph CRL SV, and ST-Graph CRL MV for the same skeleton sequences in  Figure~\ref{tsne}. Here we observe that even in this 2D embedding it is evident that the features for different classes are better separated by using ST-Graph CRLs than using P\&C FW. Points of different colors are mixed up in (a) while they are more separated in (b) and (c). Meanwhile, points of the same color in (c) are more concentrated than those in (b). For example, there is a clear line among points with different colors at the bottom right of (c) while they are mixed up at the bottom left of (b). This supports the conclusion that CRL between multi-view skeletons makes the learned representation more discriminative. 

\section{Conclusion}
In this paper, we studied the problem of view-invariant skeleton-based action recognition by learning effective representations without any manual action labeling. Based on ST-GCN structural encoder, a multi-view spatial-temporal graph contrastive representation learning approach was developed to maximize the mutual information between the representations extracted from multiple skeleton data simultaneously taken from different views. Specifically, we explored five popular skeleton data augmentation methods and found only temporal subgraph can make a positive role in multi-view CRL. Then, to support our global-local CRL, partitioning functions were designed to segment ST-Graph into multiple subgraphs along spatial or temporal dimensions and projection heads were added to map the learned representations to another latent space. Besides, we proposed a local-global spatial-temporal graph contrastive loss, combined with task uncertainty, to model the multi-scale co-occurrence relationship between spatial and temporal domains. Experiments on two multi-view action datasets showed that our proposed approach, no matter in single-view or multi-view scenarios, got competitive performance compared with the random baseline and other state-of-the-art unsupervised skeleton-based action recognition methods. 
In the future, we will explore new approaches to effectively handle multi-view multi-person scenarios.



%


%

\ifCLASSOPTIONcaptionsoff
  \newpage
\fi



%
\small
\bibliographystyle{IEEEtran}
\bibliography{MVCRL}

%
%

%








\end{document}